\documentclass[runningheads]{llncs}

\usepackage{eccv}

\usepackage{eccvabbrv}

\usepackage{graphicx}
\usepackage{booktabs}

\usepackage[accsupp]{axessibility}  

\usepackage{hyperref}

\usepackage{orcidlink}

\usepackage{xcolor}
\usepackage{multirow}
\usepackage{makecell}
\usepackage{amsmath}
\usepackage{microtype}
\usepackage{booktabs} 
\usepackage{pifont}   
\usepackage{graphicx} 
\usepackage{amssymb}
\usepackage{bbding} 

\begin{document}

\title{Rule-VLN: Bridging Perception and Compliance via Semantic Reasoning and Geometric Rectification} 

\titlerunning{Rule-VLN}


\author{
Jiawen Wen\inst{1}
\and
Penglei Sun\inst{1}(\Envelope)
\and
Wenjie Zhang\inst{1}
\and
Suixuan Qiu\inst{2}
\and
Weisheng Xu\inst{1}
\and
Xiaofei Yang\inst{3}
\and
Xiaowen Chu\inst{1}(\Envelope)
}

\authorrunning{J.~Wen et al.}

\institute{
The Hong Kong University of Science and Technology (Guangzhou), Guangzhou, China\\
\email{\{jwen341,psun012,wzhang834,wxu421\}@connect.hkust-gz.edu.cn, xwchu@hkust-gz.edu.cn}
\and
Beijing Normal University, Beijing, China\\
\email{qiusuixuan@mail.bnu.edu.cn}
\and
Guangzhou University, Guangzhou, China\\
\email{xiaofeiyang@gzhu.edu.cn}
}




\maketitle

\begin{abstract}
As embodied AI transitions to real-world deployment, the success of the Vision-and-Language Navigation (VLN) task tends to evolve from mere reachability to social compliance. However, current agents suffer from a “Goal-driven trap”, prioritizing physical geometry (“\textit{can} I go?”) over semantic rules (“\textit{may} I go?”), frequently overlooking subtle regulatory constraints. To bridge this gap, we establish Rule-VLN, the first large-scale urban benchmark for rule-compliant navigation. Spanning a massive 29k-node environment, it injects 177 diverse regulatory categories into 8k constrained nodes across four curriculum levels, challenging agents with fine-grained visual and behavioral constraints. We further propose the Semantic Navigation Rectification Module (SNRM), a universal, zero-shot module designed to equip pre-trained agents with safety awareness. SNRM integrates a coarse-to-fine visual perception VLM framework with an epistemic mental map for dynamic detour planning. Experiments demonstrate that while Rule-VLN challenges state-of-the-art models, SNRM significantly restores navigation capabilities, reducing CVR by 19.26\% and boosting TC by 5.97\%. The project page is available at \url{https://jeffry-wen.github.io/Rule-VLN/}.

\end{abstract}

\section{Introduction}
\label{sec:intro}

\begin{figure}[t]
  \begin{center}
    \centerline{\includegraphics[width=0.9\textwidth]{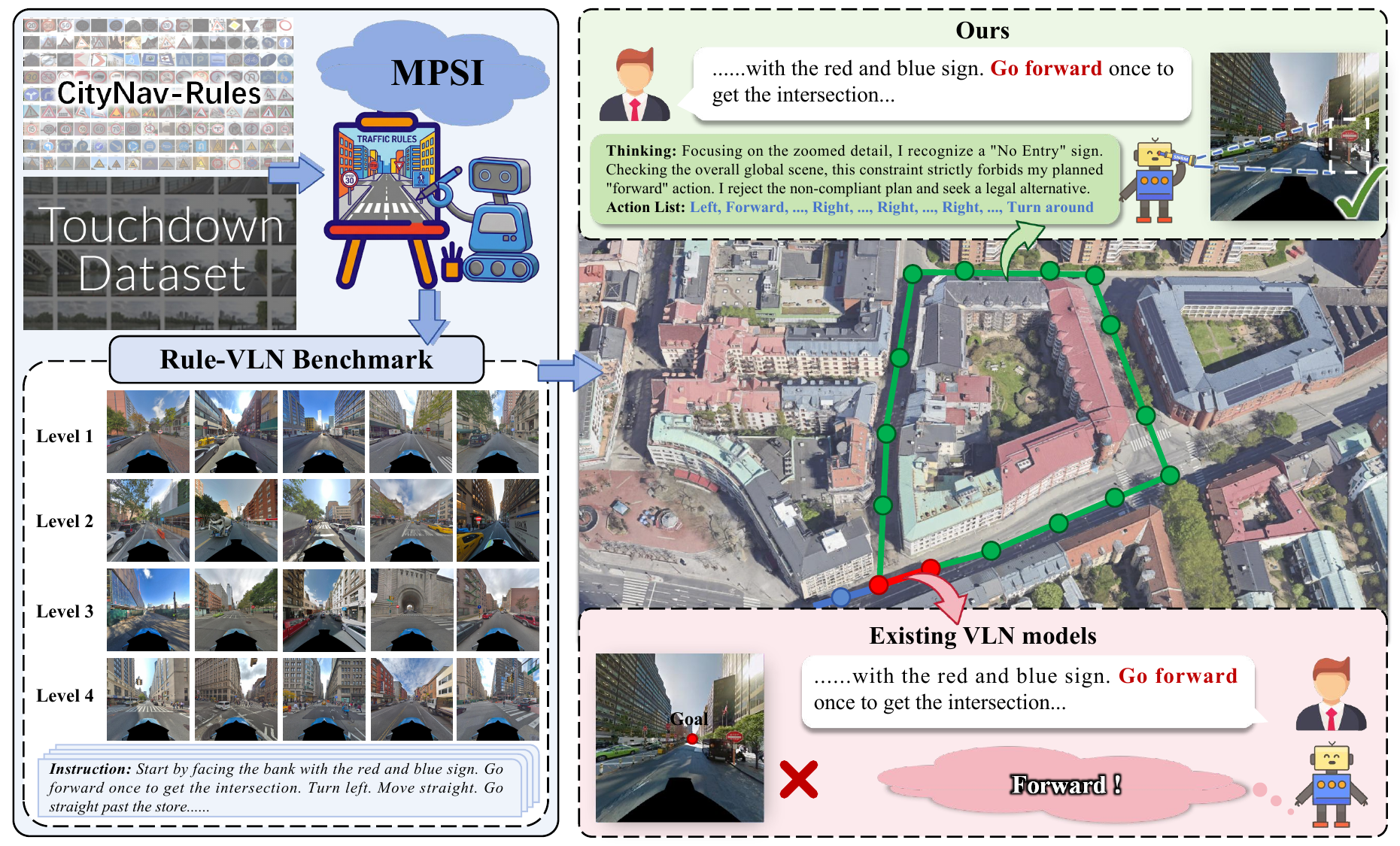}} 
    \caption{\textbf{The Rule-VLN Paradigm.} \textbf{Left:} Benchmark construction via MPSI pipeline by injecting semantic constraints into urban topologies. \textbf{Right:} Unlike standard agents (bottom) violating “No Entry” signs, our method (top) helps the agent detect prohibitions, prunes illegal actions, and executes compliant detours (green path).}
    \label{fig:Abstract_fig}
  \end{center}
\end{figure}

Vision-and-Language Navigation (VLN)~\cite{anderson2018vision} models have demonstrated superior performance across a variety of tasks. Moreover, the emergence of Multimodal Large Models (MLMs)~\cite{chen2025exploring} has injected new momentum into the VLN field. By grounding natural language instructions into visual observations, state-of-the-art agents~\cite{xiang2020learning, chen2019touchdown, sun2026terrain, anderson2018vision, zhang2024navid, chen2024mapgpt,qiao2024llm,sun20243d} demonstrate remarkable proficiency in goal-oriented planning. However, as embodied AI transitions from simulated testbeds to real-world deployment, the definition of navigation success must evolve beyond mere reachability to encompass social compliance and safety. In complex urban scenarios, valid trajectories are governed strictly by semantic rules (i.e., \textit{may} I go?) rather than just physical geometry (i.e., \textit{can} I go?)~\cite{wen2025securelargelanguagemodels,10802618}. Although physically traversable, a road may be semantically forbidden due to regulatory signage (e.g., “No Entry”); disregarding such constraints can lead to critical safety hazards. Therefore, enabling agents to perceive and follow these rule-based constraints is not just an improvement but a critical requirement for safe urban navigation and human-computer interaction~\cite{hamdani2025adaptive}.

Despite obvious progress in visual reasoning~\cite{ke2025explain}, current VLN models exhibit a critical compliance deficit. Existing works primarily focus on optimizing shortest-path efficiency or maximizing exploration coverage~\cite{10495141, chowa2026language}, often neglecting the integration of high-level regulatory signals. As shown in Figure~\ref{fig:Abstract_fig}, standard agents tend to fall into a “Goal-driven trap”, where they follow clear paths but ignore critical semantic restrictions. Existing models over-rely on salient geometry and fail to ground small-scale semantic cues effectively, preventing them from overriding geometric priors~\cite{Xiao_2025_CVPR, Li_2025_ICCV}. Furthermore, the scarcity of diverse, safety-critical training data in existing datasets~\cite{chen2019touchdown, schumann2021generating,sun2025city} exacerbates this alignment problem.

To bridge the gap between idealized paths and real-world constraints, we build upon Touchdown~\cite{chen2019touchdown} to introduce Rule-VLN. 
Compared to the existing VLN task, our dataset explicitly incorporates traffic signs, which serve as important navigational cues despite their small visual footprint.
Spanning a massive 29,000-node urban environment, Rule-VLN injects 177 diverse regulatory categories into 8,180 specific nodes systematically organized across four curriculum levels. 
To ensure these generated constraints are visually and geometrically realistic, we employ dynamic graph modification alongside a Mask-Prioritized Semantic Injection (MPSI) pipeline. Evaluations demonstrate that these non-physical obstacles significantly impede the navigation performance of current SOTA models. 
To achieve the ability of rule navigation, we propose the Semantic Navigation Rectification Module (SNRM), a plug-and-play, zero-shot module that integrates macro-to-micro visual reasoning with local dynamic mapping navigation. This training-free approach enables generic VLN models to adhere to rules, maximally increasing Task Completion (TC) by up to 5.97\% and reducing Constraint Violation Rates (CVR) by 19.26\%.

Our contributions are summarized as follows:
(1) We propose Rule-VLN, the first large-scale benchmark for rule-based urban navigation, presenting a rigorous challenge through high-quality semantic injection and dynamic graph modification. (2) We introduce SNRM, a universal, zero-shot module that equips pre-trained agents with rule compliance by effectively bridging visual perception with topological planning. (3) Extensive experiments validate the significant difficulties posed by Rule-VLN and demonstrate that SNRM effectively restores navigation capabilities and safety in constrained environments.

\section{Related Work}

\subsection{Vision-Language Navigation (VLN)}
VLN requires agents to ground natural language into long-horizon actions. In outdoor settings, benchmarks like Touchdown~\cite{chen2019touchdown} and map2seq~\cite{schumann2021generating} introduce complex graph-structured environments. Recently, foundation models have reshaped the field. NaviLLM~\cite{Zheng_2024_CVPR} introduces schema-based instruction tuning to enhance human-like reasoning, while MapGPT~\cite{chen2024mapgpt} employs map-guided prompting for adaptive global planning. NavGPT-2~\cite{zhou2024navgpt} enables large language models to have visual navigation capabilities. VELMA~\cite{schumann2024velma} verbalizes observations for reasoning, VLN-Video~\cite{li2024vln} exploits temporal video cues, and FLAME~\cite{xu2025flame} leverages synthetic data for MLLM adaptation. NavAgent~\cite{liu2024navagent} performs navigation tasks by integrating multi-scale environmental contexts, while MMCNav~\cite{zhang2025mmcnav} accomplishes its goals via a multi-agent collaborative framework.
Despite these advances, general VLMs prioritize navigational efficiency over constraint adherence, creating a granularity gap by favoring salient geometry over subtle rule signals. While retraining~\cite{chu2025sft} can address this, it requires high training costs and lowers overall navigation performance. Consequently, developing a plug-and-play module has become crucial.

\subsection{Rule-Compliant and Safe Navigation}
As VLN advances toward real-world deployment, safety has transitioned from an implicit metric to an explicit objective. 
Existing approaches primarily address safety through two lenses: physical traversability and social compliance. Methods like Safe-VLN~\cite{10496163} and others~\cite{jeong2024zero,kim2025care} focus on collision avoidance by predicting navigable areas, while VLM-Social-Nav~\cite{song2024vlm} incorporates social norms by treating them as soft cost scores during planning.
In terms of constraint modeling, recent works like CA-Nav~\cite{chen2025constraint} and GC-VLN~\cite{yin2025gc} decompose instructions into subgoals to guide heuristic search. However, a critical limitation remains: these systems typically model constraints as optimization objectives (soft rewards) rather than strict prohibitions. They struggle to handle semantic safety scenarios where a path is geometrically traversable but logically forbidden (e.g., ``No Entry'').
To address this, our SNRM framework introduces an Epistemic Mental Map that treats semantic rules as hard constraints. Unlike soft-penalty approaches, SNRM enables zero-shot, plug-and-play topological rectification, strictly enforcing rule adherence without altering the navigation backbone.

\subsection{Data Synthesis for Embodied AI}
To address the high annotation costs and long-tail disturbances in embodied AI, mechanism-oriented data synthesis has become a pivotal strategy. In VLN, specialized benchmarks like R2R-UNO~\cite{hong2024navigating}, VLN-ChEnv~\cite{liu2025vln}, and RAM~\cite{wei2025unseen} target specific failure modes ranging from path obstructions to environmental changes. General-purpose 2D editing has progressed with AnyDoor~\cite{chen2024anydoor} and SmartEdit~\cite{huang2024smartedit}. However, applying generic synthesizers to Rule-VLN presents two distinct challenges. First, standard models operating in perspective space typically ignore the equirectangular distortion inherent in panoramic views. Second, they often fail to generate verifiable regulatory signals. This lack of reliability stems from weak spatial-relation consistency~\cite{gokhale2022benchmarking, huang2025t2i}, severe text unreadability~\cite{zhang2025strict, lu2025towards}, and attribute-binding violations~\cite{kang2025counting}. Our MPSI pipeline addresses these limitations via a dual-mask conditioning strategy with explicit panoramic projection, ensuring injected rules are geometrically rectified and semantically legible.

\section{Rule-VLN Benchmark}

\subsection{Task Formulation}
\label{sec:task_formulation}

In Rule-VLN, the agent is tasked with navigating a discrete environment $G = (V, E_{\text{geo}})$ following a natural language instruction $X = (x_1, \dots, x_L)$, where $V$ denotes navigable nodes and $E_{\text{geo}}$ represents intrinsic geometric connectivity. At each step $t$, the agent at node $v_t$ perceives a panoramic observation $O_t = \{ o_{t,k} \}_{k=1}^K$, consisting of $K$ discrete visual slices. The navigation policy selects a neighbor $v_{next} \in \mathcal{N}(v_t)$ by identifying the slice $o_{t,k}$ that visually aligns with the target direction, governed by a geometric projection mapping $k = \Pi(v_t, v_{next})$.

\textbf{Semantic Connectivity vs. Geometric Connectivity.} 
Standard VLN assumes an identity mapping between geometry and traversability, \textit{i.e.}, any $e \in E_{\text{geo}}$ is navigable. Rule-VLN challenges this by introducing a Dynamic Semantic Constraint. We posit that edge traversability is conditional on rule-compliance, modeled by a binary validity mask $\mathcal{M}$:
\begin{equation}
    \mathcal{M}(v_t, v_{next} \mid X, \mathcal{R}) = \mathbb{I}\left[ \mathcal{C}(o_{t, \Pi(v_t, v_{next})} \mid \mathcal{R}) = 1 \right]
\end{equation}
where $\mathbb{I}[\cdot]$ is the indicator function and $\mathcal{C}$ evaluates whether the visual slice corresponding to the edge $e=(v_t, v_{next})$ contains regulatory prohibitions (e.g., ``Do not enter'' signs) defined in the rule set $\mathcal{R}$.

Consequently, the agent operates on a Semantically Pruned Graph $G' = (V, E_{\text{sem}})$, where $E_{\text{sem}} = \{ e \in E_{\text{geo}} \mid \mathcal{M}(e) = 1 \}$. The objective is to find a trajectory $\tau = \langle v_0, \dots, v_n \rangle$ that reaches the target defined by $X$, subject to the strict constraint that $\tau \subseteq E_{\text{sem}}$. When the intended geometric path is obstructed by $\mathcal{M}(e)=0$, the agent must suppress geometric shortest-path heuristics and infer a latent feasible detour compliant with $\mathcal{R}$.

\subsection{City Navigation Rules Dataset}
\label{subsec:dataset}

To bridge the perception-compliance gap, we introduce \textbf{CityNav-Rules-73K} (Figure~\ref{fig1:rule_vln}a), the first large-scale rule-semantic dataset explicitly coupling visual signals with fine-grained actionable constraints. Comprising 73,937 samples across 177 categories from different regions around the world, the dataset features two core innovations designed for navigation logic rather than mere classification:

\begin{figure}[ht]
  \begin{center}
    \centerline{\includegraphics[width=0.9\textwidth,height=6.5cm]{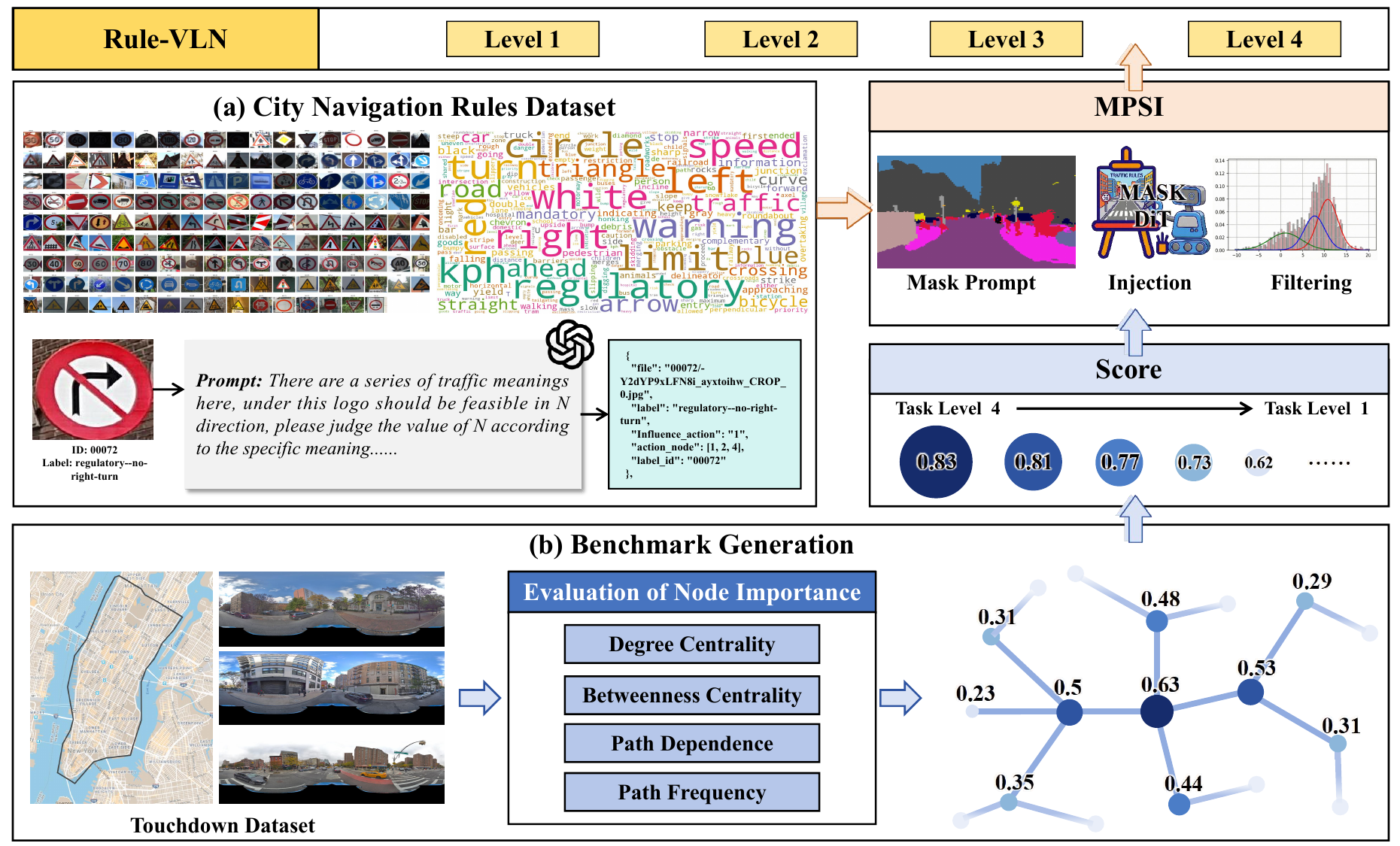}}
    \caption{\textbf{Rule-VLN Construction Pipeline.} \textbf{(a)} CityNav-Rules Dataset: Translates visual signals into permissible action constraints via LLM. \textbf{(b)} Benchmark Generation: Filters strategic nodes via topological metrics and injects constraints via MPSI to construct curriculum environments.}
    \label{fig1:rule_vln}
  \end{center}
\end{figure}

\textbf{LLM-Driven Discrete Action Mapping.} To translate abstract rules into rigorous control constraints, we employ GPT-5 to map each visual category to a Permissible Action Subspace $\mathcal{A}_{valid}$ derived from the global discrete action set (Straight, Left, Right, U-turn). This process converts semantic classifications into precise geometric lookup tables (e.g., ``no-right-turn'' yields $\mathcal{A}_{valid} = \{ \text{Straight, Left, U-turn} \}$), directly linking perception to graph traversal.

\textbf{Visual-Semantic Hybrid Representation.} We construct a Hybrid Representation interleaving Visual Descriptors (e.g., ``Red Circle'') and Semantic Imperatives (e.g., ``No Entry''). This challenges agents to decouple appearance from intent—for instance, learning that a ``Red Circle'' implies prohibition while a ``Blue Circle'' implies mandatory action. This design penalizes superficial pattern matching and enforces robust causal links between visual cues and geometric navigability.

\subsection{Touchdown-Semantic Constraint Benchmark Construction}
To strictly evaluate rule compliance, we construct Rule-VLN by injecting semantic constraints into the Touchdown graph (Fig.~\ref{fig1:rule_vln}b). We identify strategic nodes via a criticality score aggregating four metrics: (1) Degree Centrality~\cite{freeman1978centrality} for local intersection complexity; (2) Betweenness Centrality~\cite{freeman1977set} for global bottlenecks; (3) Path Dependence~\cite{albert2000error} to assess detour robustness; and (4) Path Frequency~\cite{bellemare2016unifying} to prevent shortcut memorization. Based on this score, we filter impactful nodes to establish four curriculum difficulty levels, injected via MPSI (Sec.~\ref{MPSI}). As shown in Tables~\ref{tab:benchmark_comparison} and~\ref{tab:rule_vln_stats}, constraints escalate from 31.44\% (Level-1) to 91.13\% (Level-4). Unlike limited indoor or synthetic datasets, Rule-VLN provides a real-world urban environment with 177 explicit rule types, of which a subset is injected into the node network following a 60\% action-impacting rule distribution, enforcing compliant detours through progressive difficulty levels rather than simple shortest-path heuristics.

\begin{table}[thb]
\centering
\caption{Comparison of various Vision-and-Language Navigation benchmarks in constrained environments.}
\label{tab:benchmark_comparison}
\setlength{\tabcolsep}{4pt} 
\resizebox{\textwidth}{!}{%
\begin{tabular}{@{}lccccc@{}}
\toprule
\textbf{Benchmark} & 
\textbf{Scene} & 
\begin{tabular}[c]{@{}c@{}}\textbf{Data}\\\textbf{Source}\end{tabular} & 
\begin{tabular}[c]{@{}c@{}}\textbf{Prog.}\\\textbf{Diff.}\end{tabular} & 
\begin{tabular}[c]{@{}c@{}}\textbf{Rule}\\\textbf{Cat.}\end{tabular} & 
\begin{tabular}[c]{@{}c@{}}\textbf{Path}\\\textbf{Optimization}\end{tabular} \\ \midrule
ImperfectVLN~\cite{liu2025vln} & Indoor & Matterport3D & \ding{55} & 0 & Shortest Path \\
DynamicVLN~\cite{sun2025dynamicvln} & Urban & Synthetic Simulator & \ding{55} & $<$10 & Shortest Path \\
HA-VLN 2.0~\cite{dong2025ha} & Mixed & 3D Scans + Avatars & \ding{55} & 0 (Implicit) & Shortest Path \\
Safe-VLN~\cite{10496163} & Indoor & Matterport3D & \ding{55} & 1 (Collision) & Safe Reselection \\ \midrule
\textbf{Rule-VLN (Ours)} & \textbf{Urban} & \textbf{Real-world images} & \textbf{\ding{51} (4 Levels)} & \textbf{177} & \textbf{Compliant Detour} \\ \bottomrule
\end{tabular}%
}
\end{table}

\section{Semantic Injection and Navigation Rectification Module}
\subsection{Mask-Prioritized Semantic Injection (MPSI)}
\label{MPSI}

\begin{figure}[ht]
  \begin{center}
    \centerline{\includegraphics[width=\columnwidth]{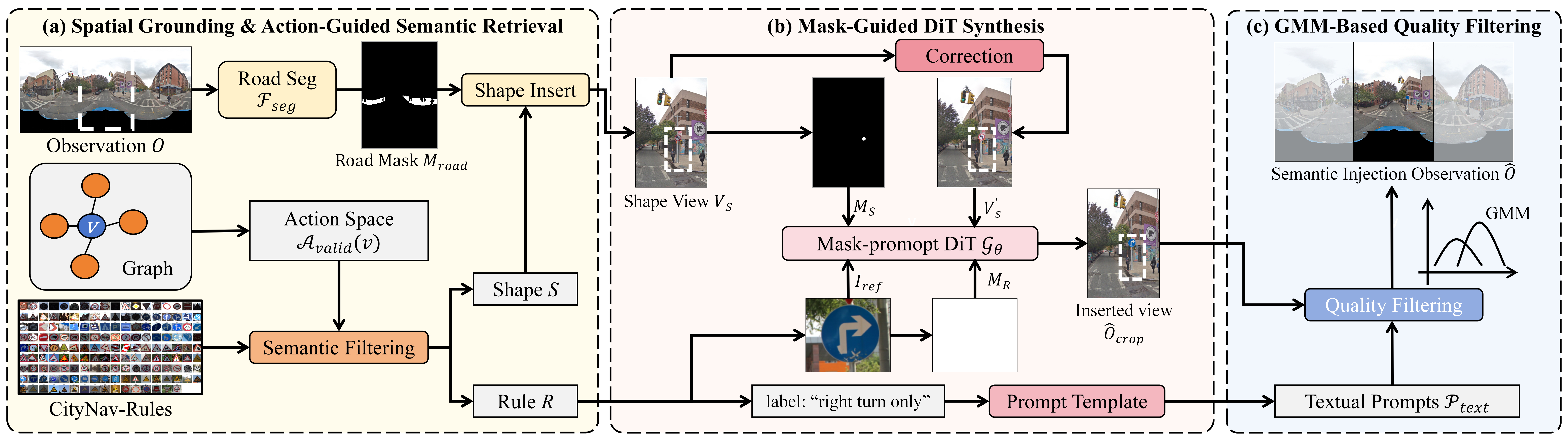}}
    \caption{\textbf{MPSI Pipeline.} \textbf{(a)} Boundary extraction via $M_{road}$ and prior retrieval. \textbf{(b)} Synthesis via dual-mask-conditioned DiT. \textbf{(c)} GMM-based filtering and stitching.}
    \label{fig1:MPSI}
  \end{center}
\end{figure}

To mitigate diffusion hallucinations, we propose MPSI (Fig.~\ref{fig1:MPSI}), a pipeline that injects node-action-compliant signs into panoramas.

\textbf{Spatial Grounding and Rule Decoupling.} We decouple regulatory signals into geometric shape $S$ and semantic rule $R$. Given a target node $v$ with permissible action subspace $\mathcal{A}_{valid}(v)$, we retrieve a semantically aligned instance $(S, R) \sim \mathcal{D}_{insert}$. To ensure global quality, we apply a cropping operator $\mathcal{C}$ to the panoramic observation $O \in \mathbb{R}^{H \times W \times 3}$, yielding $O_{crop} = \mathcal{C}(O) \in \mathbb{R}^{H_c \times W_c \times 3}$. Simultaneously, we employ a segmentation prior $\mathcal{F}_{seg}$ to extract a binary road mask $M_{road} = \mathcal{F}_{seg}(O_{crop}) \in \{0, 1\}^{H_c \times W_c}$, physically anchoring the insertion region for $S$ before semantic injection to avoid the position offset.

\textbf{Mask-Guided DiT Synthesis.}
Inspired by Insert Anything~\cite{song2025insert}, we employ a mask prompt Diffusion Transformer (DiT)~\cite{peebles2023scalable} $\mathcal{G}_{\theta}$ with dual-mask conditioning to prevent feature bleeding. First, the shape prior $S$ is projected into $O_{crop}$ via $M_{road}$ to form $V_S$, which is optically rectified into $V'_S$ to yield a precise boundary mask $M_S$. Unlike text-driven generation, we utilize a reference rule image $I_{ref}$ for visual-semantic guidance. We concatenate the binary masks ($M_S$ and rule-specific $M_R$) and spatial features $V'_S$ with the noisy latent $z_t$, while injecting semantic features $\mathcal{E}_{ref} = \text{CLIP}_{\text{img}}(I_{ref})$:
\begin{equation}
    \hat{O}_{crop} = \mathcal{G}_{\theta}\left(z_t \oplus V'_S \oplus M_S \oplus M_R, t \mid \mathcal{E}_{ref}\right)
\end{equation}
where $\oplus$ denotes channel-wise concatenation and $t$ is the diffusion timestep. This strictly confines the latent trajectory within explicit boundaries, effectively eliminating semantic hallucinations.

\textbf{GMM-Based Quality Filtering and Stitching.}
To ensure semantic legibility, we evaluate the CLIP alignment $s_{align} = \text{cos}(E_{\text{img}}(\hat{O}_{crop}), E_{\text{text}}(\mathcal{P}_{text}))$ between the synthesized crop and the rule category $\mathcal{P}_{text}$. Instead of rigid heuristics, we model the distribution of $s_{align}$ using a bimodal Gaussian Mixture Model (GMM) ($K=2$) to adaptively prune low-confidence outliers:
\begin{equation}
    p(s_{align}) = \sum_{k=1}^2 \pi_k \mathcal{N}(s_{align} \mid \mu_k, \sigma_k^2)
\end{equation}
Candidates belonging to the high-confidence mode are accepted and seamlessly blended back into the original panorama via a stitching operator $\hat{O} = \mathcal{S}(\hat{O}_{crop}, O)$, ensuring full-scale visual fidelity and geometric consistency.

\subsection{Semantic Navigation Rectification Module (SNRM)}
\label{subsec:SNRM}

\begin{figure}[ht]

  \begin{center}
    \centerline{\includegraphics[width=\textwidth,height=6.4cm]{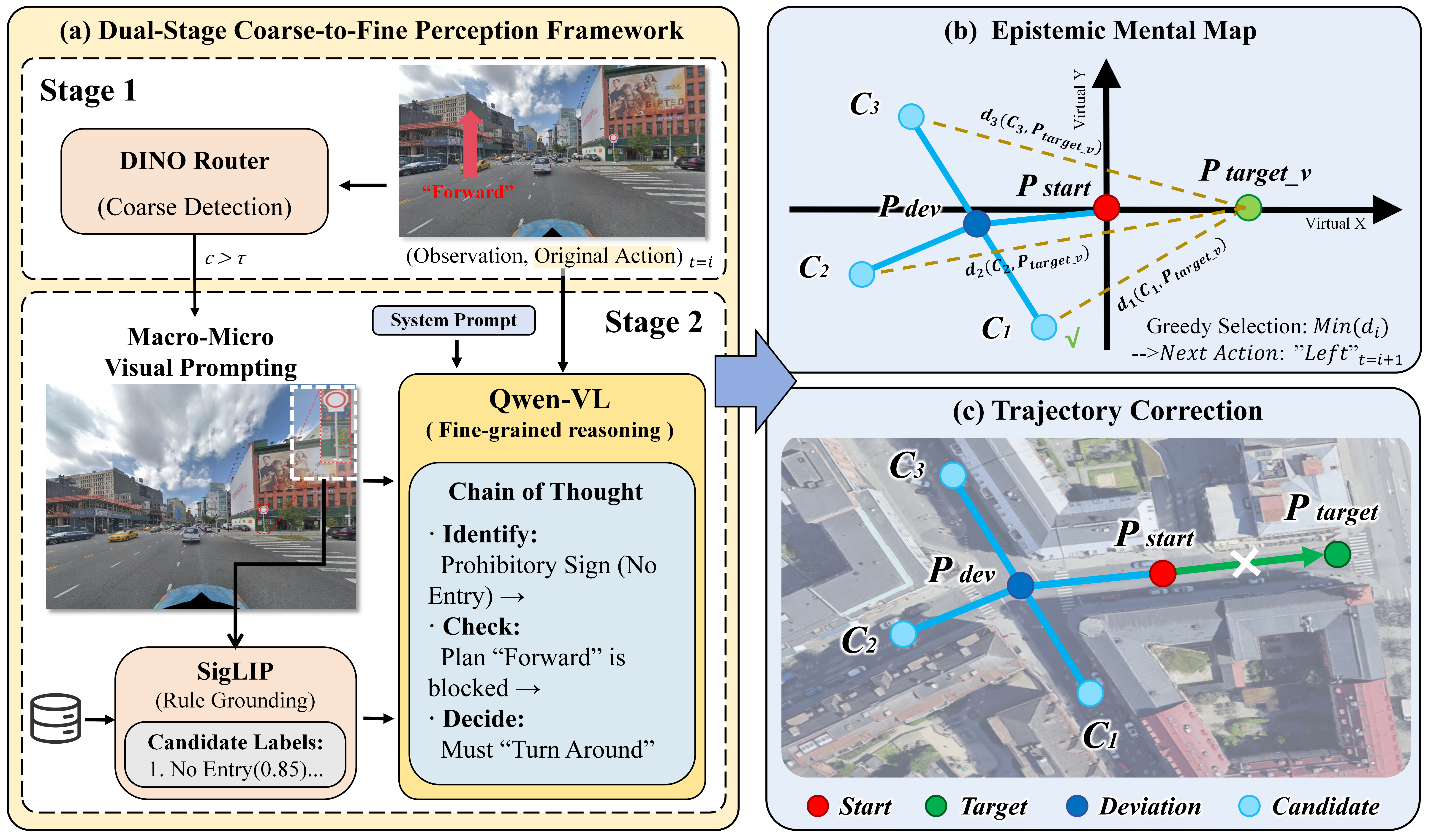}}
    \caption{\textbf{The SNRM Framework.} (a) Illustrating the dual-stage perception mechanism for rule grounding. (b-c) showing the local mental map for trajectory correction.}
    \label{fig:SNRM}
  \end{center}
\end{figure}

To bridge the granularity gap, we propose SNRM, a plug-and-play, model-agnostic module that acts as a semantic reflection module for any frozen VLN policy $\pi_{base}$. SNRM decomposes compliance into a dual-stage perception framework (Fig.~\ref{fig:SNRM}).

\textbf{Dual-Stage Coarse-to-Fine Perception Framework.}
To reliably extract subtle regulatory cues from complex observations, SNRM employs a coarse-to-fine pipeline (Figure \ref{fig:SNRM} a). It can be divided into two main stages. \textbf{\textit{Stage 1: Macro-Micro Visual Prompting.}} Motivated by visual prompting~\cite{wu2024visual}, we design a human-like attention method that follows a “global-to-local” perception paradigm. To mitigate computational costs, a lightweight detector (e.g., DINO) first scans the panorama $O_t$ for potential cues using generic prompts $P_{detect}$. Upon detection, we generate a Macro-Micro Visual Prompt: a global view $V_{macro}$ with highlighted bounding boxes to preserve topological context, and a magnified crop $V_{micro}$ to enhance fine-grained symbol resolution. By incorporating explicit visual prompts, this approach enhances the model's perception of fine-grained rule signals while maintaining robust global awareness. This ensures that the generated actions are both rule-compliant and consistent with environmental connectivity. \textbf{\textit{Stage 2: Knowledge-Driven Rule Grounding (KDRG).}} To prevent open-ended hallucinations, we anchor VLM inference to a predefined regulatory knowledge base $\mathcal{K}$, a lightweight textual rule-name bank containing the 177 normalized CityNav-Rules categories. We employ SigLIP to compute the zero-shot similarity between $V_{micro}$ and rule descriptions in $\mathcal{K}$, retrieving the top-ranked text prior $T_{rule} = \arg\max_{t \in \mathcal{K}} \text{Sim}(V_{micro}, t)$. Finally, the tuple $(V_{macro}, V_{micro}, T_{rule})$ and the intended action $a_{base} \sim \pi_{base}$ are fed into a VLM (e.g., Qwen-3VL). Through structured Chain-of-Thought (CoT) reasoning, the VLM outputs a safety token $s \in \{\text{Safe, <Correct Action>}\}$, strictly gating the execution of $a_{base}$.

\textbf{Epistemic Mental Map and Trajectory Correction.}
Upon detecting a conflict, SNRM overrides the base policy via a dynamic 2D virtual topology (Fig.~\ref{fig:SNRM} b-c). Defining the conflict origin as $P_{start}=(0,0)$, we anchor the intended target at $P_{target\_v} = (\sin(\Delta \theta_{exp}), \cos(\Delta \theta_{exp}))$ based on the expected heading $\Delta \theta_{exp}$. As the agent executes forced compliant actions to a deviation point $P_{dev}$ (updated via dead reckoning), SNRM replans by evaluating candidates $\mathcal{C} = \{C_1, \dots, C_n\}$ through a penalized greedy heuristic:
\begin{equation}
   C_{best} = \arg\min_{C_i \in \mathcal{C}} \Big( \| P(C_i) - P_{target\_v} \|_2 + \lambda \cdot \mathbb{I}_{backtrack}(P(C_i)) \Big)
\end{equation}
where $P(C_i)$ is the predicted coordinate, and $\mathbb{I}_{backtrack}$ applies a severe penalty $\lambda$ if $P(C_i)$ falls within a critical radius of visited nodes. To further robustify against spatial drift, we employ a visual memory buffer: if the cosine similarity between current and start node features $\text{cos}(f_{curr}, f_{start}) > \tau_{sim}$, a closed-loop trap is detected, triggering a forced detour correction.

\section{Experiment}
\subsection{Experimental Setup}

\textbf{Evaluation Metrics.}
We evaluate navigation via Task Completion (TC)~\cite{chen2019touchdown}, Shortest-Path Distance (SPD)~\cite{chen2019touchdown} and Success weighted by Path Length (SPL)~\cite{anderson2018evaluation}. To rigorously quantify rule adherence, we introduce the Constraint Violation Rate (CVR), which normalizes the number of violations $V_i$ by the agent's actual exposure to active constraints ($\mathcal{R}_{neighbor}$):
\begin{equation}
    \text{CVR} = \frac{1}{N} \sum_{i=1}^{N} \frac{V_i}{\sum_{t=1}^{L_i} \mathbb{I}(\text{Node}_t \in \mathcal{R}_{neighbor})}
\end{equation}
where $L_i$ is path length and the indicator function $\mathbb{I}[\cdot]$ identifies steps with proximal regulatory signals. Synthesis quality is measured via PSNR~\cite{wang2004image}, SSIM~\cite{wang2004image}, LPIPS~\cite{zhang2018unreasonable}, and FID~\cite{heusel2017gans}.

\textbf{Implementation Details.}
We benchmark against ORAR~\cite{schumann2022analyzing}, Loc4plan~\cite{tian2024loc4plan}, VELMA~\cite{schumann2024velma}, and FLAME~\cite{xu2025flame} using official hyperparameters. To ensure rule visibility, we retrained models that take partial views as input, such as FLAME, by turning the FOV from $60^{\circ}$ to $120^{\circ}$ and fine-tuning. The MPSI was performed on a single NVIDIA RTX 4090 GPU. Training model is conducted on 4$\times$ NVIDIA H100 GPUs, while inference and SNRM deployment utilize a single H100. Notably, our proposed SNRM is training-free.

\begin{figure}[thb]

  \begin{center}
\centerline{\includegraphics[width=\textwidth,height=5cm]{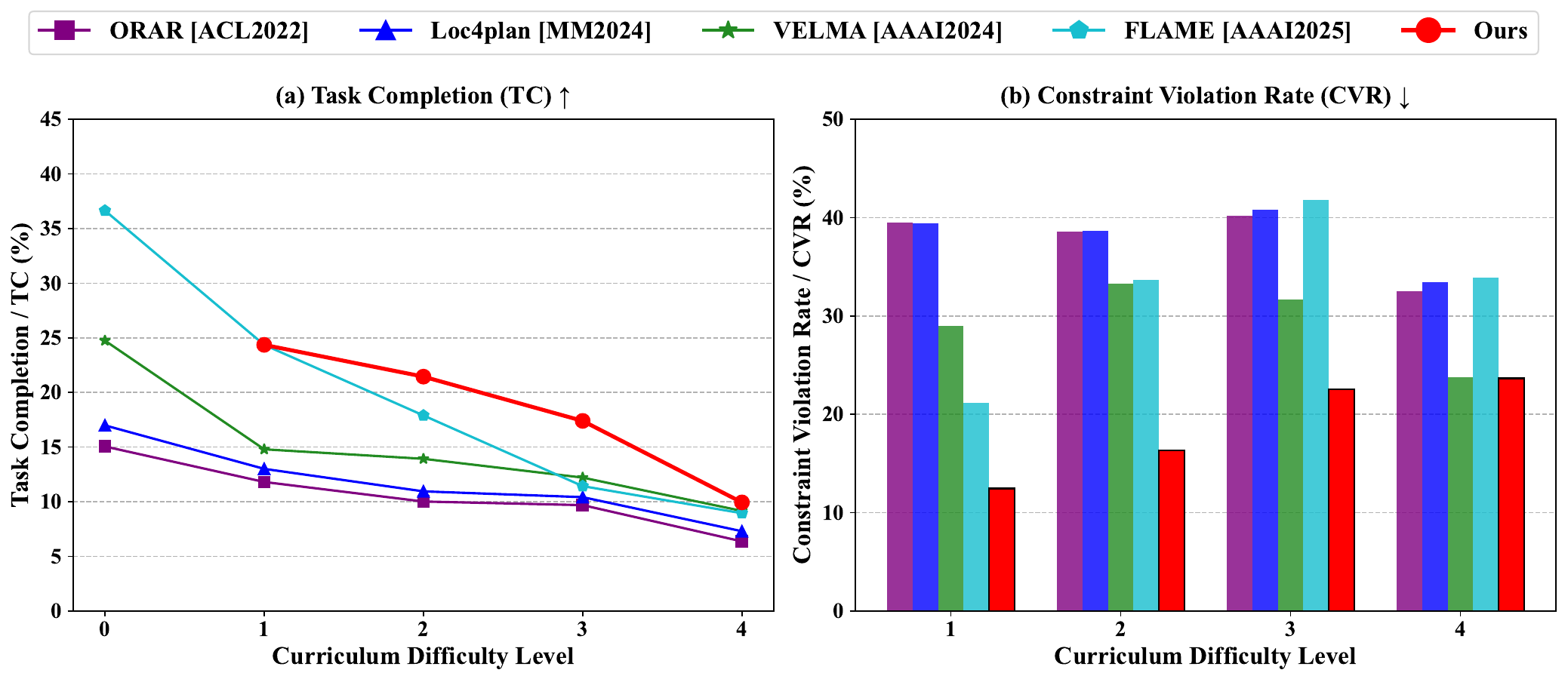}}
    \caption{Performance metrics of SOTA models on the Rule-VLN benchmark.}
    \label{fig:rule_vln_performance}
  \end{center}
\end{figure}

\subsection{Evaluating the Robustness of Current VLN Methods in Rule-VLN}
\label{sec:rule_vln_challenge}
\textbf{Analysis of Evaluation Results.} 
The algorithmic injection of regulatory symbols establishes a rigorous four-level curriculum (Table~\ref{tab:rule_vln_stats}). To establish baselines, we evaluate state-of-the-art (SOTA) VLN architectures. As illustrated in Fig.~\ref{fig:rule_vln_performance}, the introduction of explicit semantic constraints induces a catastrophic performance collapse across all paradigms, compared to an environment without constraints (Level 0). Even recent foundation models like FLAME exhibit a sharp degradation in Task Completion (TC) alongside alarmingly high Constraint Violation Rates (CVR, often exceeding 30-40\%).

\begin{figure}[bht]
  \begin{center}
    \centerline{\includegraphics[width=\columnwidth,height=6.5cm]{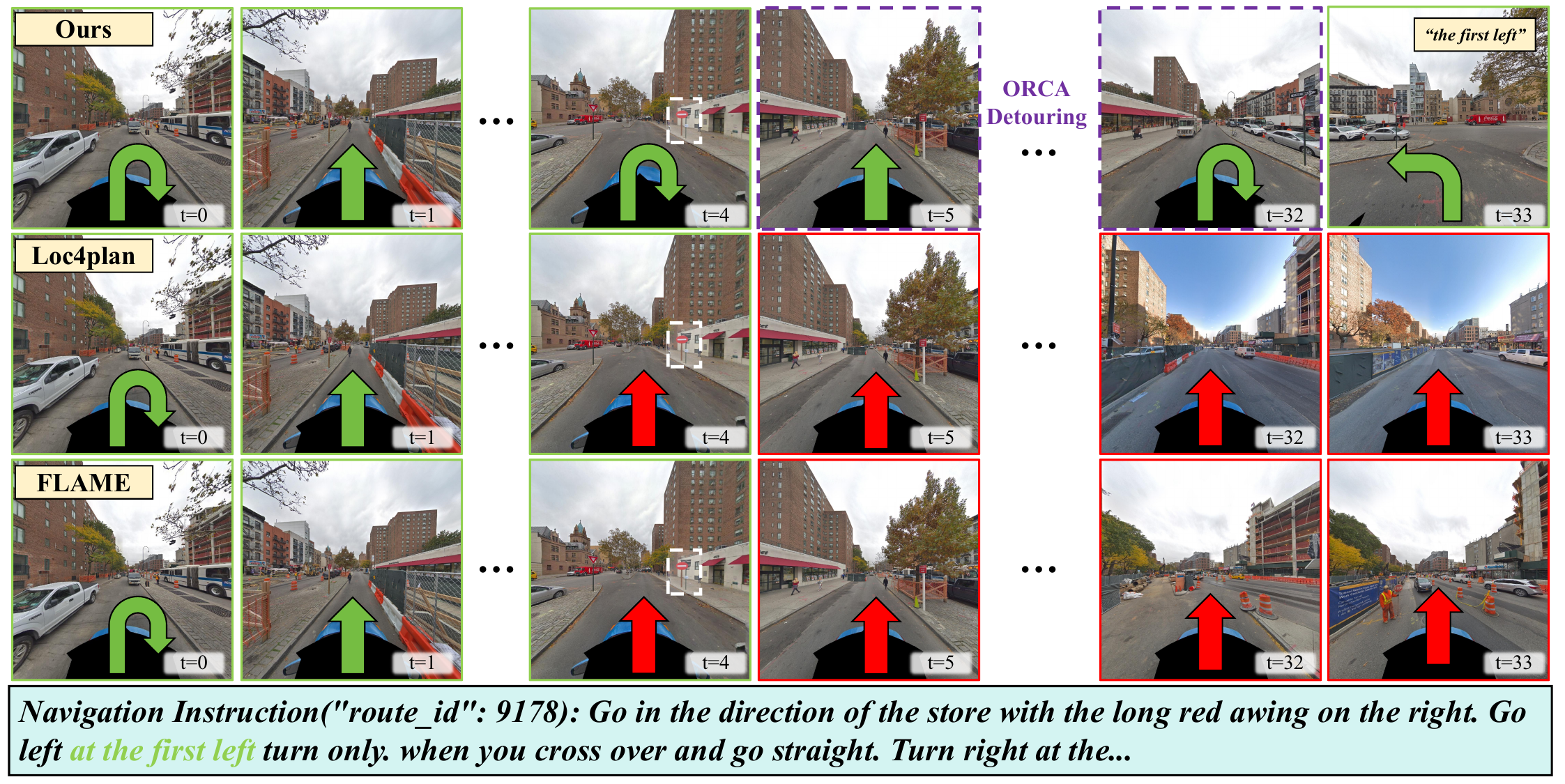}}
    \caption{Visualization results of our method and other baselines on navigation samples. Green arrows indicate strictly compliant and correct actions, while red arrows indicate semantic violations or navigation failures.}
    \label{fig:traj_vis}
  \end{center}
\end{figure}

This phenomenon highlights a severe granularity gap: current models overfit to idealized topologies and remain entirely blind to fine-grained semantic imperatives (e.g., driving into ``No Entry'' zones). These baseline results validate Rule-VLN as a necessary testbed and forcefully underscore the need for our proposed SNRM module, which we analyze quantitatively in Sec.~\ref{Rule_Compliance_SNRM}.

\textbf{Trajectory Visualization Analysis.} 
Figure~\ref{fig:traj_vis} visualizes a scenario where a prohibitory sign obstructs the path at $t=4$. While baselines (Loc4plan, FLAME) exhibit semantic blindness by executing illegal forward actions, SNRM successfully grounds the visual cue and triggers a preemptive U-turn. Leveraging its epistemic mental map, the agent executes a legal detour to bypass the restricted zone, ultimately realigning with the global instruction (``the first left'') at $t=33$. This visually corroborates SNRM's ability to transform blind geometric traversal into robust, rule-compliant navigation.

\subsection{Evaluation of the SNRM}
\label{Rule_Compliance_SNRM}
Table~\ref{tab:main_results} demonstrates SNRM's zero-shot integration with Loc4plan and FLAME across four difficulty levels. Table~\ref{tab:efficiency_comparison} compares the real-time inference performance after incorporating the SNRM module.

\textbf{Universal Plug-and-Play Efficacy.} SNRM bridges the granularity gap across diverse architectures, yielding significant gains for both traditional (Loc4plan) and MLLM-based (FLAME) agents. For Loc4plan (Level-2), SNRM not only reduces CVR by 8.70\% but also improves TC by 1.53\%, proving its effectiveness even without heavy reasoning priors. This impact is amplified in FLAME (Level-3), where SNRM slashes CVR by 19.26\% while boosting TC by 5.97\%, validating its capability to transform blind traversal into semantically constrained navigation regardless of the backbone. Although SNRM inevitably leads to more detours during navigation, it still improves SPL in most experimental settings.

\textbf{Robustness and Compliance.} In Level-1, despite identical TC (24.36\%), SNRM improves efficiency (SPD: 16.58 vs. 17.12) and strictly enforces compliance (CVR: 12.47\% vs. 21.17\%). As constraints intensify in Level-3, the baseline collapses (TC 11.43\%, CVR 41.79\%), whereas SNRM exhibits superior robustness, recovering TC to 17.40\% while suppressing violations. This advantage persists in Level-4, validating SNRM as a reliable safety envelope for complex navigation.


\textbf{Inference latency and navigation performance.}
As shown in Table~\ref{tab:efficiency_comparison}, incorporating the VLM introduces additional inference latency, but the event-triggered design of SNRM limits this overhead by relying on the base navigation model in most cases. For SNRM-8B, the trigger rate is only about 26\%; taking FLAME as an example, the average per-step latency increases from 0.095~s to 1.158~s, while the CVR decreases by 18.25 percentage points. To further improve practical efficiency, we design a 4B SNRM variant without CoT, which adds only 0.224~s latency per step while still reducing CVR by 10.55 percentage points.

\begin{table}[thb]
    \centering
    \caption{Performance comparison across different tasks on the Rule-VLN benchmark. The performance differences brought by integrating SNRM are indicated in subscripts ($\uparrow$ for increases and $\downarrow$ for decreases). Best results within each category are highlighted in \textbf{bold}.}
    \label{tab:main_results}
    \scriptsize
    \setlength{\tabcolsep}{3.5pt}
    \begin{tabular}{ll ccc ccc}
        \toprule
        \multirow{2}{*}{Level} & \multirow{2}{*}{Metric} & \multicolumn{3}{c}{Traditional Models} & \multicolumn{3}{c}{VLM/LLM-based Models} \\
        \cmidrule(lr){3-5} \cmidrule(lr){6-8}
        & & ORAR\cite{schumann2022analyzing} & Loc4plan\cite{tian2024loc4plan} & \textbf{+SNRM} & VELMA\cite{schumann2024velma} & FLAME\cite{xu2025flame} & \textbf{+SNRM} \\
        \midrule
        \multirow{4}{*}{Level-1}
        & TC $\uparrow$    & 11.81 & 13.01 & \textbf{14.07}$_{\textcolor{gray}{\uparrow 1.06}}$ & 14.80 & \textbf{24.36} & \textbf{24.36}$_{\textcolor{gray}{- 0.00}}$ \\
        & SPL $\uparrow$   & 7.42  & 7.72  & \textbf{7.76}$_{\textcolor{gray}{\uparrow 0.04}}$  & 9.87  & \textbf{18.83} & 17.72$_{\textcolor{gray}{\downarrow 1.11}}$ \\
        & SPD $\downarrow$ & 21.72 & 21.81 & \textbf{21.31}$_{\textcolor{gray}{\downarrow 0.50}}$ & 18.52 & 17.12 & \textbf{16.58}$_{\textcolor{gray}{\downarrow 0.54}}$ \\
        & CVR $\downarrow$ & 39.48 & 39.45 & \textbf{36.48}$_{\textcolor{gray}{\downarrow 2.97}}$ & 29.01 & 21.17 & \textbf{12.47}$_{\textcolor{gray}{\downarrow 8.70}}$ \\
        \midrule
        \multirow{4}{*}{Level-2}
        & TC $\uparrow$    & 10.02 & 10.95 & \textbf{12.48}$_{\textcolor{gray}{\uparrow 1.53}}$ & 13.93 & 17.90 & \textbf{21.45}$_{\textcolor{gray}{\uparrow 3.55}}$ \\
        & SPL $\uparrow$   & 6.63  & 6.78  & \textbf{6.98}$_{\textcolor{gray}{\uparrow 0.20}}$  & 9.59  & \textbf{13.94} & 13.91$_{\textcolor{gray}{\downarrow 0.03}}$ \\
        & SPD $\downarrow$ & 23.04 & 23.54 & \textbf{23.03}$_{\textcolor{gray}{\downarrow 0.51}}$ & 19.06 & 20.04 & \textbf{18.30}$_{\textcolor{gray}{\downarrow 1.74}}$ \\
        & CVR $\downarrow$ & 38.55 & 38.67 & \textbf{29.97}$_{\textcolor{gray}{\downarrow 8.70}}$ & 33.30 & 33.68 & \textbf{16.35}$_{\textcolor{gray}{\downarrow 17.33}}$ \\
        \midrule
        \multirow{4}{*}{Level-3}
        & TC $\uparrow$    & 9.69  & 10.42 & \textbf{12.61}$_{\textcolor{gray}{\uparrow 2.19}}$ & 12.21 & 11.43 & \textbf{17.40}$_{\textcolor{gray}{\uparrow 5.97}}$ \\
        & SPL $\uparrow$   & 6.54  & 6.43  & \textbf{7.16}$_{\textcolor{gray}{\uparrow 0.73}}$  & 7.84  & 9.03  & \textbf{10.51}$_{\textcolor{gray}{\uparrow 1.48}}$ \\
        & SPD $\downarrow$ & 22.71 & 23.39 & \textbf{22.23}$_{\textcolor{gray}{\downarrow 1.16}}$ & 20.08 & 23.44 & \textbf{19.96}$_{\textcolor{gray}{\downarrow 3.48}}$ \\
        & CVR $\downarrow$ & 40.22 & 40.80 & \textbf{33.64}$_{\textcolor{gray}{\downarrow 7.16}}$ & 31.69 & 41.79 & \textbf{22.53}$_{\textcolor{gray}{\downarrow 19.26}}$ \\
        \midrule
        \multirow{4}{*}{Level-4}
        & TC $\uparrow$    & 6.37  & 7.30  & \textbf{7.96}$_{\textcolor{gray}{\uparrow 0.66}}$  & 9.16  & 8.95  & \textbf{9.94}$_{\textcolor{gray}{\uparrow 0.99}}$ \\
        & SPL $\uparrow$   & 4.10  & 4.43  & \textbf{4.54}$_{\textcolor{gray}{\uparrow 0.11}}$  & 5.87  & 6.24  & \textbf{6.39}$_{\textcolor{gray}{\uparrow 0.15}}$ \\
        & SPD $\downarrow$ & 23.95 & 24.50 & \textbf{23.63}$_{\textcolor{gray}{\downarrow 0.87}}$ & 21.18 & 26.29 & \textbf{25.61}$_{\textcolor{gray}{\downarrow 0.68}}$ \\
        & CVR $\downarrow$ & 32.50 & 33.42 & \textbf{27.00}$_{\textcolor{gray}{\downarrow 6.42}}$ & 23.77 & 33.89 & \textbf{23.68}$_{\textcolor{gray}{\downarrow 10.21}}$ \\
        \bottomrule
    \end{tabular}
\end{table}

\begin{table}[t]

\caption{Efficiency and performance comparison.}
\label{tab:efficiency_comparison}

\scriptsize
\resizebox{\columnwidth}{!}{
\begin{tabular}{lccccc}
\toprule
Method
& CVR$\downarrow$
& \makecell[c]{Trigger Rate$\downarrow$}
& \makecell[c]{Avg. Inf.\\Time/Step$\downarrow$}
& \makecell[c]{Rule-triggered\\Step Time$\downarrow$}
& \makecell[c]{No-rule\\Step Time$\downarrow$} \\
\midrule

Loc4plan-11M
& 31.40
& --
& \textbf{0.002 s}
& --
& -- \\

+SNRM-8B (CoT)
& \textbf{21.00}
& 26.36\%
& 1.132 s
& 1.541 s
& 0.148 s \\

\midrule

FLAME-8B
& 36.86
& --
& \textbf{0.095 s}
& --
& -- \\

+SNRM-8B (CoT)
& \textbf{18.61}
& 26.25\%
& 1.158 s
& 3.852 s
& \textbf{0.217 s} \\

+SNRM-4B
& 26.31
& \textbf{14.49\%}
& 0.319 s
& \textbf{0.941 s}
& 0.234 s \\

\bottomrule
\end{tabular}
}
\end{table}

\subsection{Ablation experiments and analysis of SNRM}
To dissect the contribution of each module within SNRM, we conduct a comprehensive ablation study on the most challenging Level-4 configuration (Table~\ref{tab:ablation_SNRM}). The baseline setup (\#1), devoid of SNRM's components, exhibits poor navigation success (TC: 8.95) and a high rule violation rate (CVR: 33.89\%), acting as a naive geometric navigator.

\textbf{Impact of the Dual-Stage Perception Framework.} 
The removal of either Macro-Micro Visual Prompting(MMVP) (Setup \#3) or Knowledge-Driven Rule Grounding(KDRG) (Setup \#2) explicitly degrades performance. Without MMVP (\#3), the VLM struggles to simultaneously focus on subtle regulatory symbols and the global topological context, leading to an increased CVR (26.86\%). Similarly, removing the KDRG module (\#2) deprives the VLM of deterministic text priors, causing it to occasionally hallucinate or misclassify long-tail signs, which marginally drops the TC. Setup \#5 demonstrates that these two perception modules act synergistically to reliably translate raw visual cues into actionable semantic constraints.

\textbf{Impact of Topological Recovery.} 
Setup \#4 (w/o Mental Map) achieves the lowest CVR (21.28\%) but compromises TC (6.53\%), as the agent stops at constraints without a path-generation mechanism. Conversely, the full framework (\#5) improves TC to 9.94\% and SPD to 25.61, demonstrating that the Epistemic Mental Map is essential for translating static rule recognition into active navigation and detour computing.

\begin{table}[htb]
    \centering
    
    
    \begin{minipage}[t]{0.48\textwidth}
        \centering
        \caption{Statistics of affected nodes and instructions across Rule-VLN levels.}
        \label{tab:rule_vln_stats}
        \resizebox{\linewidth}{!}{%
            \begin{tabular}{@{} lccc @{}}
                \toprule
                Level & \# Nodes & Aff. Node(\%) & Aff. Instr.(\%) \\
                \midrule
                1 & 1941 & 9.99  & 31.44 \\
                2 & 2029 & 10.45 & 54.79 \\
                3 & 2163 & 11.14 & 74.52 \\
                4 & 2047 & 10.54 & 91.13 \\
                \midrule
                Total   & 8180 & 42.13 & 99.93 \\
                \bottomrule
            \end{tabular}%
        }
    \end{minipage}
    \hfill 
    \begin{minipage}[t]{0.48\textwidth}
        \centering
        \caption{Ablation study of SNRM components on Level-4.}
        \label{tab:ablation_SNRM}
        \resizebox{\linewidth}{!}{ 
            \begin{tabular}{@{} c ccc ccc @{}}
                \toprule
                \multirow{2}{*}{Setup} & \multicolumn{3}{c}{SNRM Components} & \multicolumn{3}{c}{Metrics} \\
                \cmidrule(lr){2-4} \cmidrule(lr){5-7}
                & MMVP & KDRG & Map & TC ($\uparrow$) & SPD ($\downarrow$) & CVR ($\downarrow$) \\
                \midrule
                \#1 & $\times$    & $\times$    & $\times$    & 8.95 & 26.29 & 33.89 \\
                \#2 & \checkmark  & $\times$    & \checkmark  & 9.02 & 25.65 & 25.94 \\
                \#3 & $\times$    & \checkmark  & \checkmark  & 8.88 & 25.69 & 26.86 \\
                \#4 & \checkmark  & \checkmark  & $\times$    & 6.53 & 26.91 & \textbf{21.28} \\
                \textbf{\#5} & \checkmark  & \checkmark  & \checkmark  & \textbf{9.94} & \textbf{25.61} & 23.68 \\
                \bottomrule
            \end{tabular}
        }
    \end{minipage}
    
    \vspace{0.6cm} 
    
    
    \begin{minipage}[t]{0.48\textwidth}
        \centering
        \caption{Synthesis quality and efficiency comparison.}
        \label{tab:image_quality}
        \resizebox{\linewidth}{!}{%
            \begin{tabular}{@{} l ccccc @{}}
                \toprule
                \multirow{2}{*}{Method} & \multicolumn{2}{c}{Fidelity} & \multicolumn{2}{c}{Perceptual} & \\
                \cmidrule(lr){2-3} \cmidrule(lr){4-5}
                & PSNR$\uparrow$ & SSIM$\uparrow$ & LPIPS$\downarrow$ & FID$\downarrow$ & \multirow{-2}{*}{VRAM(G)$\downarrow$} \\
                \midrule
                FLUX.1 & 24.59 & 0.7518 & 0.3096 & 30.31 & 36.71 \\
                \textbf{MPSI} & \textbf{32.41} & \textbf{0.9280} & \textbf{0.0503} & \textbf{7.72} & \textbf{15.12} \\
                \bottomrule
            \end{tabular}%
        }
    \end{minipage}
    \hfill 
    \begin{minipage}[t]{0.48\textwidth}
        \centering
        \caption{Performance comparison of different VLMs within SNRM on Level-4.}
        \label{tab:vlm_comparison}
        \resizebox{\linewidth}{!}{
            \begin{tabular}{@{} l c ccc @{}}
                \toprule
                Method & Param & TC ($\uparrow$) & SPD ($\downarrow$) & CVR ($\downarrow$) \\
                \midrule
                FLAME + SNRM$_{\text{Qwen3-vl}}$ & 8B  & \textbf{9.94} & \textbf{25.61} & 23.68 \\
                FLAME + SNRM$_{\text{Qwen3-vl}}$ & 4B  & 8.88 & 25.67 & 24.10 \\
                FLAME + SNRM$_{\text{LLaVA}}$ & 7B  & 7.74 & 28.10 & \textbf{21.17} \\
                FLAME + SNRM$_{\text{Intern}}$ & 8B  & 9.45 & 25.79 & 23.45 \\
                FLAME + SNRM$_{\text{Pixtral}}$ & 12B & 8.38 & 26.43 & 29.74 \\
                \bottomrule
            \end{tabular}
        }
    \end{minipage}
    
\end{table}

\textbf{Impact of VLM Backbones.}
We instantiate SNRM with Qwen3-Vl~\cite{bai2025qwen3}, InternVL-3.5~\cite{wang2025internvl3}, LLaVA-1.6~\cite{liu2024llavanext}, and Pixtral~\cite{agrawal2024pixtral} to explore the influence of different VLM models on the perception of internal rules. Intra-family scaling (Qwen-VL) confirms that parameter capacity enhances spatial CoT execution. Qwen-VL and InternVL achieve the optimal balance via robust grounding. In contrast, LLaVA-1.6 is overly conservative; its low CVR (21.17\%) leads to excessive detouring and navigation failure (low TC). Finally, Pixtral 12B’s underperformance against 8B models confirms that for embodied tasks, visual-spatial grounding is more critical than parameter scale.
\begin{figure}[!ht]
  \begin{center}
    \centerline{\includegraphics[width=\columnwidth]{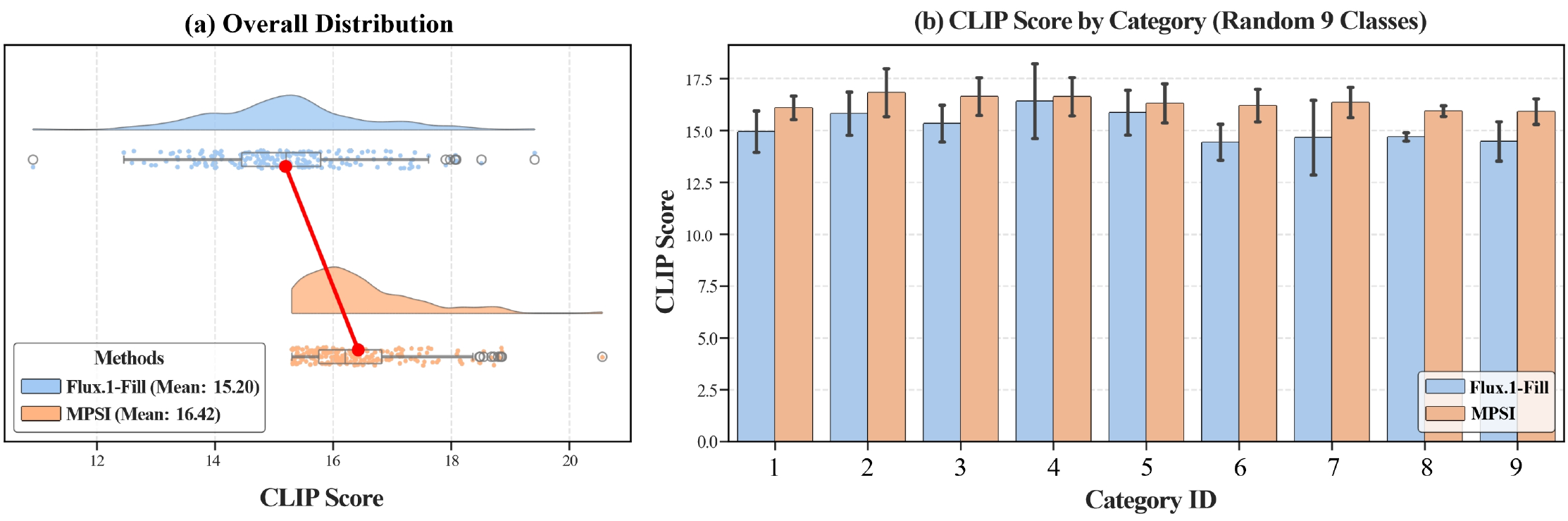}}
    \caption{Quantitative evaluation of semantic alignment using CLIP scores. (a) Overall score distribution. (b) Distribution across randomly sampled categories.}
    \label{fig:clip_score}
  \end{center}
\end{figure}

\subsection{High-Fidelity Semantic Injection via MPSI}
\textbf{Visual Realism and Fidelity.} As shown in Table~\ref{tab:image_quality}, MPSI significantly outperforms the FLUX.1-Fill~\cite{labs2025flux} baseline. High pixel-level fidelity (SSIM 0.9280, PSNR 32.41) confirms that our dual-mask conditioning strictly confines the latent space and prevents artifact bleeding into the native panorama. Furthermore, MPSI drastically improves perceptual realism (FID 7.72, LPIPS 0.0503), effectively mitigating the geometric distortions typical of generic diffusion models in localized synthesis. Due to the masking design, MPSI achieves a 58.81\% reduction in GPU memory footprint compared to FLUX.1-Fill.

\textbf{Semantic Alignment and Filtering.} To ensure the injected signs serve as actionable constraints, we evaluate cross-modal semantic legibility via CLIP scores (Fig.~\ref{fig:clip_score}). MPSI achieves a higher overall mean (16.42 vs. 15.20). Crucially, the raincloud plot (Fig.~\ref{fig:clip_score}a) reveals that while FLUX exhibits a long tail of low-scoring outliers (indicating hallucinations), MPSI's distribution is sharply truncated at the lower end. This explicitly validates our GMM-Based Quality Filtering in actively pruning corrupted generations. Moreover, MPSI demonstrates consistent superiority across diverse rule categories (Fig.~\ref{fig:clip_score}b), guaranteeing unambiguous visual cues for downstream tasks.

\begin{figure}[htb]

  \begin{center}
    \centerline{\includegraphics[width=\columnwidth]{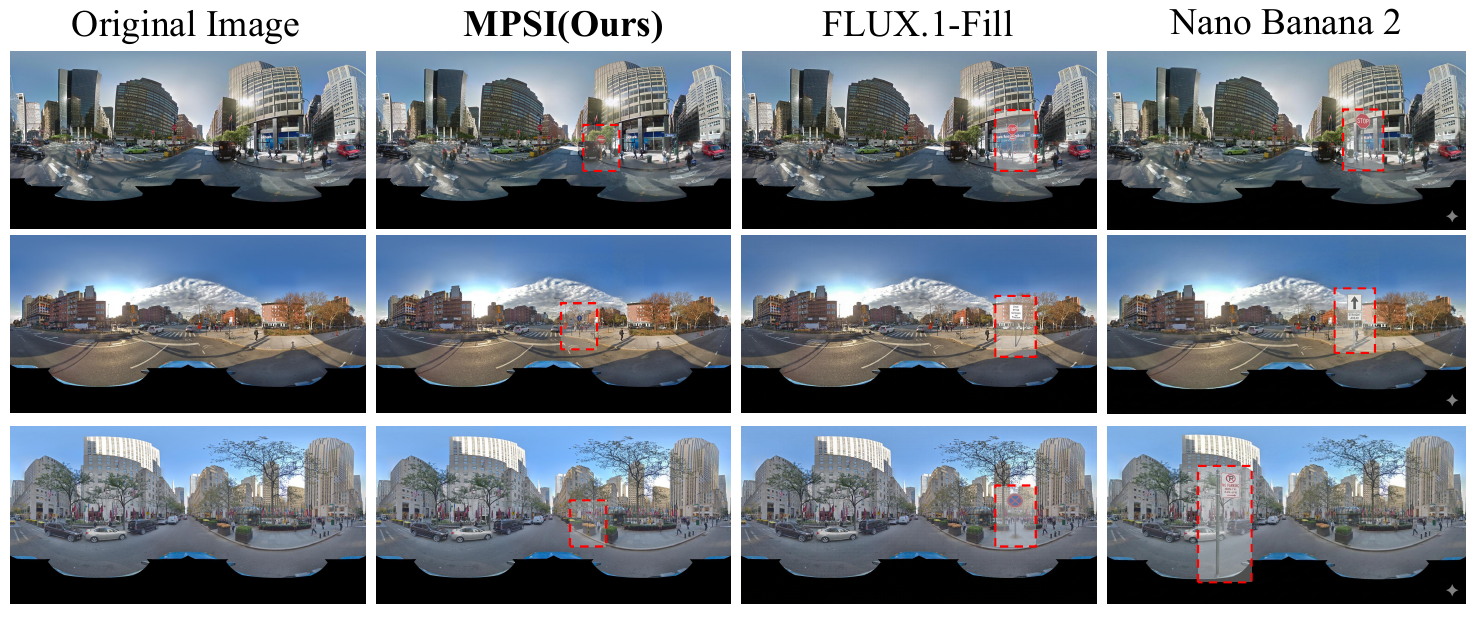}}
    \caption{Qualitative analysis of image inpainting results from MPSI, FLUX.1-Fill, and Google Nano Banana 2 across diverse urban scenarios.}
    \label{fig:MPSI_vis}
  \end{center}
\end{figure}

\textbf{Qualitative Comparison.} Fig.~\ref{fig:MPSI_vis} compares MPSI, FLUX.1-Fill, and Google Nano Banana 2. FLUX suffers from severe semantic hallucinations, and Nano Banana 2 yields spatially misplaced injections due to a lack of positional conditioning. MPSI consistently achieves superior visual fidelity and precise geometric grounding.

\section{Limitations and Future Work}
Despite its effectiveness, our framework has several limitations. On the data synthesis side, although MPSI injects rules under dual-mask constraints, the quality of the synthesized data remains affected by the complexity of the environment and the quality of rule-related images. Moreover, following the discrete-node task formulation of Touchdown may lead to viewpoint-consistency issues when the same rule is observed from adjacent nodes. For SNRM, we design its components, such as QwenVL and SigLIP, in a plug-and-play manner to enhance generalization to diverse real-world regulations. This enables SNRM to benefit from advances in foundation models without retraining the entire module, but it also introduces additional inference latency. Future work will explore stronger generative vision models and 3D geometric constraints to narrow the gap between synthesized and real-world scenes. We will also investigate model distillation, feature caching, and quantization to reduce computational overhead while preserving generalization.

\section{Conclusion}
In this work, we introduce semantic rule constraints into Vision-and-Language Navigation (VLN) to address the issue where existing agents prioritize geometric accessibility over social compliance in real-world environments. We propose Rule-VLN, the first benchmark incorporating such rule compliance challenges into VLN. This is achieved via a novel Mask-Prioritized Semantic Injection (MPSI) pipeline that precisely integrates fine-grained traffic rule signals into urban topology and visual observations. Leveraging Rule-VLN, we demonstrate the limitations of current VLN methods and further introduce the Semantic Navigation Rectification Module (SNRM). This module enables agents to effectively adapt to rule-constrained environments through macro-micro visual prompting and an epistemic mental map method. Our approach achieves state-of-the-art results on Rule-VLN without requiring alterations to the original model architectures. We believe that addressing the shift from physical accessibility to social compliance is critical for the practical deployment of VLN agents and for evaluating their capacity for safe navigation under complex rule constraints.


\section*{Acknowledgements}
We thank the reviewers for their valuable comments and constructive suggestions. We also gratefully acknowledge the computing resources provided by The Hong Kong University of Science and Technology (Guangzhou) through the Red Bird MPhil Program.

%
%
\bibliographystyle{splncs04}
\bibliography{main}
\end{document}